\documentclass{article}

\usepackage[preprint]{neurips_2024}

\usepackage[utf8]{inputenc} %
\usepackage[T1]{fontenc}    %
\usepackage{hyperref}       %
\usepackage{url}            %
\usepackage{booktabs}       %
\usepackage{amsfonts}       %
\usepackage{nicefrac}       %
\usepackage{microtype}      %
\usepackage{xcolor}         %
\usepackage{comment}

\usepackage{graphicx}

\title{Library Learning Doesn't: The Curious Case of the Single-Use ``Library''}

\author{%
  Ian Berlot-Attwell\\
  University of Toronto\\
  Vector Institute \\
  \texttt{ianberlot@cs.toronto.edu} \\
  \And
  Frank Rudzicz \\
  Dalhousie University \\
  Vector Institute \\
  \texttt{frank@dal.ca} \\
  \AND
  Xujie Si \\
  University of Toronto \\
  Vector Institute \\
  \texttt{six@cs.toronto.edu} \\
}

\begin{document}

\newcommand{\rvs}{\textit{request db}}
\newcommand{\prover}{\textsc{Prover}}
\newcommand{\evolver}{\textsc{Evolver}}
\newcommand{\lp}{LEGO-Prover}
\newcommand{\trove}{TroVE}

\newcommand{\tskip}{\textsc{Skip}}
\newcommand{\tcreate}{\textsc{Create}}
\newcommand{\timport}{\textsc{Import}}

\maketitle

\begin{abstract}
 Advances in Large Language Models (LLMs) have spurred a wave of LLM library learning systems for mathematical reasoning. %
 These systems aim to learn a reusable library of \textit{tools}, such as formal Isabelle lemmas \citep{DBLP:books/sp/Paulson94} or Python programs that are tailored to a family of tasks. Many of these systems are inspired by the human structuring of knowledge into reusable and extendable concepts \citep{DBLP:conf/pldi/EllisWNSMHCST21}, but do current methods actually learn reusable libraries of tools?

  We study two library learning systems for mathematics which both reported increased accuracy: %
  \lp\ \citep{lego} and \trove\ \citep{trove}. We find that function reuse is extremely infrequent on miniF2F \citep{minif2f} and MATH \citep{hendrycksmath2021}. %
  Our followup ablation experiments suggest that, rather than reuse, self-correction and self-consistency are the primary drivers of the observed
  performance gains. Our code and data are available at \url{https://github.com/ikb-a/curious-case}. 
\end{abstract}

\section{Introduction}

Mathematical progress is made by building with, and building upon, the tools of those who came before. Consequently, it is no surprise that there is research interest in  developing systems that can automatically learn such reusable mathematical tools. Recently, LLMs have enabled new tool-learning methods with improved performance \citep{lego, trove, agentOpt, craft} -- but are these systems truly learning generalized, reusable knowledge or is  performance improved through other mechanisms? In this work, we study two prior systems: \lp\ which aims to learn reusable formal Isabelle lemmas, and \trove\ which aims to learn reusable Python functions. For both, our analysis of the model's behaviour reveals that direct reuse is negligible. Furthermore, we perform two ablation studies supporting our position that function reuse plays a limited role in these systems' improved mathematical reasoning.

\section{Related Work}

LLM library learning, i.e., creating and reusing tools, depends on LLMs' ability to use tools. Prior evaluations of tool-use (typically assuming tools as REST APIs)  \citep{DBLP:journals/corr/abs-2405-17935}  included real-world queries \citep{berkeley-function-calling-leaderboard}, dedicated test environments \citep{DBLP:conf/emnlp/LiZ000YLHL23}, and metrics ranging from LLM-as-a-judge \citep{DBLP:conf/acl/GuoCWLQLL0L24} to tracking task-checkpoint completion \citep{DBLP:journals/corr/abs-2408-04682}.

In contrast, the evaluation of library learning systems has been limited. Accuracy is the metric of choice \citep{lego, trove, agentOpt, craft}, but cannot capture the extent or quality of reuse: an excellent library is useless to a weak reasoner, and a powerful reasoner can ignore a useless library and derive results from first principles. Prior attempts to evaluate library learning have been limited to static measures of individual functions such as cyclomatic complexity \citep{1702388,agentOpt} and abstract syntax tree depth \citep{trove}, or have answered specific questions such as the ease of human verification \citep{trove}, accuracy under domain transfer \citep{agentOpt, creator}, or performance in the sub-problem of refactoring ground truth solutions\citep{DBLP:conf/naacl/LinCHYLLL24}.

In this study, we evaluate two library learning systems for mathematical reasoning: \lp, and \trove\ (see Sections \ref{sec:lp_overview} and \ref{sec:trove_overview}). For a review of library learning systems, see Appendix \ref{rel_work_ext}.

\subsection{\lp: Purpose \& Architecture} \label{sec:lp_overview}

\lp\ consumes a set of proposed theorems %
to produce corresponding formal Isabelle \citep{DBLP:books/sp/Paulson94} proofs. It was evaluated on the miniF2F \citep{minif2f} dataset: each problem was attempted 100 times, and the system obtained feedback from the Isabelle verifier after each attempt. \lp\ was designed to perform library learning. Using the term \textit{skills} in place of \textit{tools}, \cite{lego} claimed that ``\lp\ enables LLMs to utilize existing skills retrieved from the library'' and  ``[m]odular and reusable skills are constantly added to the library to enable tackling increasingly intricate mathematical problems.'' \lp\ performs library learning via two LLM systems: 1) The \prover\ which uses the library to create proofs, and 2) the \evolver\ which iteratively refines the library. They communicate through shared databases, such as the \rvs\ which stores proposed lemmas to be proven and added to library.

\subsection{\trove: Purpose \& Architecture}\label{sec:trove_overview}

\trove\ is a ``method for inducing a toolbox of reusable functions to use in solving programmatic tasks,'' designed to receive a stream of word problems without a ground truth or verifier \citep{trove}. For each problem, it attempts to produce a Python program that prints the correct solution. \trove's mathematical reasoning was evaluated with the MATH dataset \cite{hendrycksmath2021}. Each problem is considered once: an LLM generates 15 solutions, and the best is selected based on self-consistency (i.e., majority vote)  \citep{DBLP:conf/iclr/0002WSLCNCZ23}. In generation, 5 solutions ignore the library and directly generate a program (\tskip\ mode), 5 create a reusable helper function for inclusion in the library (\tcreate\ mode), and 5 use a function from the library (\timport\ mode).

\section{Analysis of \lp}\label{sec:lp}

\begin{table}
  \caption{Lemma reuse in \lp\ released logs. Note that \textbf{lemma reuse is very uncommon}, and  \textbf{no lemma reused twice}. %
  For each split, we report the number of problems solved, the number of unique lemmas occurring in the \prover's input prompts, the number of lemmas reused verbatim once, or more than once, and the number of lemmas whose \textit{name} is reused once, or more than once. A lemma is reused $N$ times if it appears in $N+1$ solutions (i.e., the initial use, and then $N$ reuses).}
  \label{tab:lp_reuse}
  \centering
  \begin{tabular}{lrrllll}
    \toprule
    \multicolumn{3}{c}{} & \multicolumn{2}{c}{Verbatim reused} & \multicolumn{2}{c}{Name reused} \\
    \cmidrule(r){4-5}
    \cmidrule(r){6-7}
    Split     & Problems Solved     & Lemmas in Prompts & 1 & 2+ & 1 & 2+ \\
    \midrule
    valid+GPT & 127  & 374  & 0 & 0 & 1 & 0     \\
    valid+Human & 135 & 265 & 0 & 0 & 1 & 0     \\
    test+GPT & 111 & 255 & 0 & 0 & 2 & 0     \\
    test+Human & 122 & 339 & 1 & 0 & 2 & 0 \\
    \bottomrule
  \end{tabular}
\end{table}

We begin by analyzing the publicly released \lp\ evaluation log files \footnote{\url{https://github.com/wiio12/LEGO-Prover/blob/357672c7751cd0c84aff6bf72a3d1bf97614e81d/result/lego_result.zip}}  \citep{lego}.
These logs are a subset of the unreleased \prover\ logs corresponding to the final attempts on the successfully solved problems. Note that \lp\ was evaluated on 4 data splits, and learned over 20,000 lemmas overall \citep{lego}. 

We find that only 1,233 lemmas ($\sim$6\%) are used in the final solving step (i.e., are inputs to the \prover). Of these, exactly one lemma is reused by the \prover, and it is reused once (i.e., appears verbatim in two solutions). As the \prover\ may be adjusting a lemma (e.g., paraphrasing, commenting, etc...) we repeat the analysis, checking only for the lemma's name. Again, lemma reuse is rare, and no lemma is reused more than once (i.e., no lemma has its \textit{name} appear in 3 or more solutions). See Table \ref{tab:lp_reuse} for details. For an example of verbatim vs. name use, see Appendix \ref{app:reuse_examples}.

Given these findings, there are only two possibilities by which \lp\ may be performing reuse: 1) indirect reuse (e.g., the learned tools are useful, reusable exemplars, rather than directly used in the final solution), or 2) direct reuse occurs in the \evolver. 

Instead, we hypothesize that reuse is not significantly boosting performance. We propose that self-correction \citep{DBLP:journals/corr/abs-2308-03188} via the \rvs\ is the main mechanism of action. %
Note that the \prover\ populates the \rvs\ by: 1) adding lemmas that the LLM suggests may be helpful sub-steps, and 2) adding lemmas from solution attempts that Isabelle could not verify. The \evolver\ uses the \rvs\ to modify existing tools to ``aid in solving requests'', and to ``resolv[e] decomposed sub-goals'' using the library \citep{lego}. Thus, the performance gains may be due to a combination of chain-of-thought \citep{DBLP:conf/nips/Wei0SBIXCLZ22} (through the \prover's proposal of helpful lemmas for the \evolver\ to solve) and self-correction (through the \evolver's retrying of failed lemmas).

\begin{figure}
  \centering
  \includegraphics[width=0.75\linewidth]{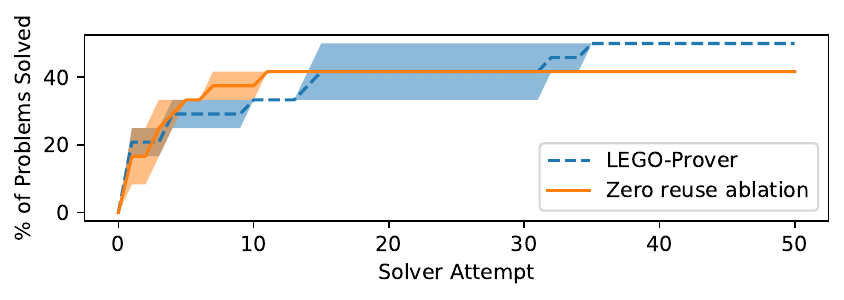}
  \caption{\lp\ performance on a subset of the miniF2F validation split. The ablated model %
  cannot reuse lemmas and performs similarly. The shaded region is one standard deviation, capturing variations in LLM output and race conditions. %
  } \label{fig:lp_ablat}
\end{figure}

To test whether any form of reuse is increasing performance, we ablate \lp\ to remove cross-problem sharing: each theorem is solved with its own independent state and databases. E.g., in place of a global \rvs, each problem now has its own independent \rvs. %
We evaluate on a random size 12 subset of the validation split and use 50 attempts per problem.  We perform our ablation using OpenAI's GPT-4o-mini as the original results were published using now deprecated versions of GPT-3.5-Turbo; see Appendix \ref{app:hyper_lp} for full details of the ablation. Running 2 trials, we find that the ablation's performance is strong, solving only 1 question less than the baseline (see Figure \ref{fig:lp_ablat}). Studying the problems solved by only the baseline, we find that only the simplest of the input lemmas are possibly used (namely $a^2 \geq 0$ and $ax^2 +bx+c = 0 \Rightarrow c=-(ax^2 +bx)$; see Appendix \ref{app:lp_baseline_proofs}). It is unclear as these facts are not treated as lemmas, and are given different justifications. This suggests that: 1) the LLM may be too weak if it needs examples of basic facts 2) the LLM struggles at reuse as it does not copy the given, verified, proofs.

\section{Analysis of \trove}\label{sec:trove}

As \trove\ logs were not released, we re-ran \trove\ on MATH, achieving accuracy within $\pm$2\% (absolute) of reported (see Appendix, Table \ref{app:tab_trove_reprod}). Note that the \trove\ library also learns import statements; we ignore these in our analysis for two reasons. Firstly, our interest is in whether the system learns and reuses non-trivial tools, unlike statements such as ``\texttt{import math}'' and ``\texttt{from sympy import symbols}''. %
Secondly, as \trove\ includes the entire library as part of the \timport\ prompt, and import statements are innately simple, it is impossible to determine whether an import statement is included in the LLM output due to reuse, or the LLM's innate knowledge.

Analyzing the logs, we find that \trove's final libraries only contain 15 learned functions, having learned functions for only 3 of the 7 MATH subject test splits: counting, number, and pre-algebra. No functions are learned in the algebra, geometry, intermediate algebra, or pre-calculus splits. Of the 15 learned functions, only 2  are reused in a correct solution: \texttt{is\_perfect\_square(n)} is reused in one correct solution %
and \texttt{is\_prime(num)} is reused in two correct solutions. %

Given 3 successful reuses in 3,201 test questions, we believe that \trove's improvements over the baselines are not due to function reuse. Instead, we believe that ensembling and self-consistency are responsible. To test this, we ablate the model by disabling \timport\ mode, but maintaining the 15 solution attempts: we generate 8 solutions ignoring the library (i.e., \tskip\ mode) and 7 attempting to create a helper function (i.e., \tcreate\ mode). As in the original work we use \texttt{CodeLlama-7b-Instruct-hf} \citep{DBLP:journals/corr/abs-2308-12950}; see Appendix \ref{app:hyper_trove} for the full ablation details. Ablating \timport\ mode prevents reuse as the library never appears in the model's input, thus also preventing library learning of import statements. As to why this ablation could still be performant, prior work  established the benefits of self-consistency and increased sampling \citep{DBLP:journals/corr/abs-2407-21787}, and it's known that library-less tool-creation can boost performance by forcing abstract reasoning \citep{craft}. 

We evaluate our ablated model on the intermediate\_algebra test split (reportedly the largest performance gain over non-reuse baselines), and the geometry, number, and count test splits. On the intermediate\_algebra, number, and count splits, our ablation exceeds the baseline's performance, with the improvement being statistically significant on two splits (See Table \ref{tab:trove_ablat_details}). %
On only the geometry split does the base model perform slightly better, though the learned libraries only contains import statements. From this we can conclude that library learning  \textit{import statements} can be slightly beneficial, but only for certain domains. Typically, \trove's library learning degrades its performance.

\begin{table}
  \caption{\trove\ performance on MATH for the ablation and the baseline. Mean and standard deviation over 5 trials are reported. The variations arise from LLM output. $\dagger$ indicates that mean ablation performance is significantly strictly higher than the baseline's, at the Bonferroni-corrected 0.05 level, using a 2-sample 1-sided Welch's t-test (note, this test assumes approximate normality).}
  \label{tab:trove_ablat_details}
  \centering
  \begin{tabular}{lllll}
    \toprule
    & \multicolumn{4}{c}{Accuracy on MATH test split}                   \\
    \cmidrule(r){2-5}
    Model     & count     & geo &  inte & num \\
    \midrule
    \trove\ Reproduced & 0.236 $\pm$ 0.008 & \textbf{0.058} $\pm$ 0.004 & 0.120 $\pm$ 0.006 & 0.258 $\pm$ 0.007    \\
    No Reuse Ablation & \textbf{0.250} $\pm$ 0.000$\dagger$ & 0.050 $\pm$ 0.000 & \textbf{0.134} $\pm$ 0.014 & \textbf{0.290} $\pm$ 0.014$\dagger$  \\
    \bottomrule
  \end{tabular}
\end{table}

\section{Conclusions}

In this study, we find that both \trove\ and \lp\ do not directly reuse the tools they learn. Furthermore, the results of our ablations suggest that their performance gains cannot be solely attributed to indirect reuse either.

We intend that this paper be a call for the better understanding of the limitations of current library learning systems, and for improved evaluation. We show that accuracy is misleading in isolation: the system's reuse behaviour is paramount, and careful ablation is critical. Both papers studied made sensible claims as the created systems were deliberately designed for library learning and were tested against ablations that were not unreasonable -- however they also relied heavily on accuracy as a metric instead of directly observing the systems' use of the library, and both chose ablations that in hindsight were too aggressive. It is clear that, particularly for ablations of library learning systems, minimal changes are preferable, and considerable thought should be put into other possible causes of improvements. There is a clear need for a broadly applicable framework for the evaluation of library learning specifically; this framework must rely on more than task accuracy and ablations to evaluate library learning and reuse.

Finally, considering library learning for mathematics in general: are LLMs capable learning tools and performing direct, verbatim reuse? Given that the observed improvements do not come from direct reuse, would direct reuse actually improve systems for mathematical reasoning, or is it overly brittle making soft reuse desirable? These important questions follow from our findings, and should inform the design of future research into library learning systems.

\section{Limitations \& Broader Impact}\label{sec:lims}

Due to resource constraints, our ablation studies could be more thorough. Most obviously, we only study two models, and on two datasets. %
The \lp\ ablation is not ideal, as library learning is disadvantaged by operating on a subset of the questions; %
this was necessary due to resource constraints.
Another limitation is that \lp's databases are pre-loaded with the full dataset of problems; consequently, the \evolver s are exposed to other problem statements -- note, however, that the impact on testing reuse is minimal. Firstly, the \prover\ cannot attempt to solve any of these other problems, thus the \rvs\ cannot gain pending lemmas related to other problems. Secondly, under the ablated model, tasks cannot share lemmas -- any performance gains would come from having access to other sample problems instead of reuse.

While we demonstrate that the performance gains in mathematical reasoning seen by \trove\ and \lp\ cannot be attributed to the direct learning and reuse of tools, there is a very important but \textit{subtly different} question which remains unanswered: whether these systems are at all capable of library learning. It is possible that these systems have the capacity to learn reusable functions and lemmas, but the datasets do not provide the opportunity. Manually inspecting the MATH dataset, our tentative conclusion is that the dataset is intrinsically not amenable to function learning with Python -- we suspect the questions are too diverse, with the shared components already being captured by standard libraries. How this could be more formally demonstrated remains an important open question that is beyond the scope of this work.

This work has no immediate societal impact, rather, it  highlights current limitations and challenges assumptions in this field. However, deploying tool-learning systems may carry a security risk from executing LLM-generated code (we sandboxed \trove). More generally, library learning systems are self-improving through code generation, an approach that has raised concerns \citep{DBLP:journals/corr/abs-2310-02304}. Unexpected behaviours may develop, thus requiring sandboxing and  monitoring, at the very least.

\begin{ack}
Resources used in preparing this research were provided, in part, by the Province of Ontario, the Government of Canada through CIFAR, and companies sponsoring the Vector
Institute \url{www.vectorinstitute.ai/partnerships/}.
Generous support was also provided by the Microsoft Accelerating Foundation Models Research (AFMR) program.

We would also like to thank Zhiruo Wang, Zhaoyu Li, William Cunningham, and our anonymous reviewers for their time and conversations that helped in various ways to shape and improve this work.
Finally, the lead author would like to thank Frank Rudzicz for years of guidance and support, and Xujie Si for both encouraging this work as being of interest to the mathematical reasoning community, and for providing critical resources without which it could not have been possible. Thank you everyone for helping make this work possible.
\end{ack}

\bibliography{sample}

\appendix

\section*{Appendix}

\section{Extended Related Work}\label{rel_work_ext}

Current LLM-based library learning systems tend to fall into two main camps: systems designed for general word problem solving, typically including mathematical reasoning and typically generating Python functions (e.g., \citet{DBLP:conf/iclr/Cai00CZ24, craft, trove}), and agentic systems designed to interact with a specific, complex environment (e.g., \citet{DBLP:journals/tmlr/WangX0MXZFA24, tan2024cradleempoweringfoundationagents, DBLP:journals/corr/abs-2402-07456, agentOpt, DBLP:journals/corr/abs-2402-15809}). 

Generally, such systems access the library via in-context learning (ICL); some place the entire library in the context \citep{trove, agentOpt}, whereas others first use a semantic-similarity retrieval step to allow for larger libraries. \citet{craft} in particular uses a retrieval system that incorporates a LLM-generated description of the tool to be retrieved; \lp\ behaves similarly by having several phases where the system alternates between proposing useful tools to be added to the library, attempting to create these tools, and possibly retrieving these tools. 

These systems are typically bottom-up (iteratively developing a library over time), though a handful of top-down approaches exist. These top-down approaches instead decompose a high-level description of the tasks into reusable modules \citep{DBLP:conf/iclr/ChenSLHG24, DBLP:journals/corr/abs-2407-09886, DBLP:journals/corr/abs-2402-15809, DBLP:conf/icassp/ZhangHLYWL24}; to the best of the authors' knowledge this approach is yet to be applied to mathematical reasoning.

These LLM-based systems typically attempt to produce reusable tools via ICL: prompting the LLM to generate ``reusable functions''. In comparison, an older family of library learning work (e.g., Dreamcoder \citep{DBLP:conf/pldi/EllisWNSMHCST21} and LILO \citep{DBLP:conf/iclr/GrandWBOLTA24}) instead frame library learning as a matter of compression. In principle a function that compresses a set of solutions must be broadly applicable, and in practice a high-level function reduces the symbolic search space for program induction. More generally, compression has been of long standing interest in the field of artificial intelligence. \citet{DBLP:journals/aim/Rendell83} defined conceptual knowledge as the ability to compress a raw space of possibilities into useful classes, and there are long-standing connections between compression and inductive reasoning. Framing inductive reasoning as the task of capturing the underlying pattern in a provided substring for the purposes of prediction,  \citet{DBLP:journals/iandc/Solomonoff64a} formalized induction as Bayesian reasoning under a prior favouring low Kolmogorov complexity. In other words, formalizing the concept of Occam's razor -- that the simplest solution, that which can be highly compressed into a short description, is more likely. For a recent treatise on the value of compression, specifically within the area of mathematical reasoning, see \citet{DBLP:journals/corr/abs-2403-04571}.

Turing our attention to mathematics, deep learning in general and LLMs in particular have found broad application in theorem proving \citep{DBLP:journals/corr/abs-2404-09939}. Considering library learning specifically, a very closely related branch of work considers the problem of refactoring a collection of ground-truth solutions into reusable components. ATG \citep{DBLP:conf/naacl/LinCHYLLL24} and REFACTOR \citep{DBLP:conf/iclr/ZhouWLG24} train models to extract reusable formal lemmas from a provided set of ground-truth formal proofs. Similarly, ReGAL \citep{DBLP:conf/icml/Stengel-EskinPB24} refactors ground-truth Python solutions for the MATH dataset into a reusable library. These systems are valuable and may represent a better first step towards reusable knowledge, but their dependence on ground-truth solutions prevents them from being conventional library learning systems. In comparison, \lp\ attempts to learn reusable lemmas and produce formal proofs from only formal problem statements, and informal natural language proofs -- furthermore, \citet{lego} demonstrated that the latter could be automatically generated by ChatGPT with only a small degradation in system performance.

\section{Example of Verbatim Use versus Name Use by \lp} \label{app:reuse_examples}

\begin{figure}
  \centering
  \includegraphics[width=\linewidth]{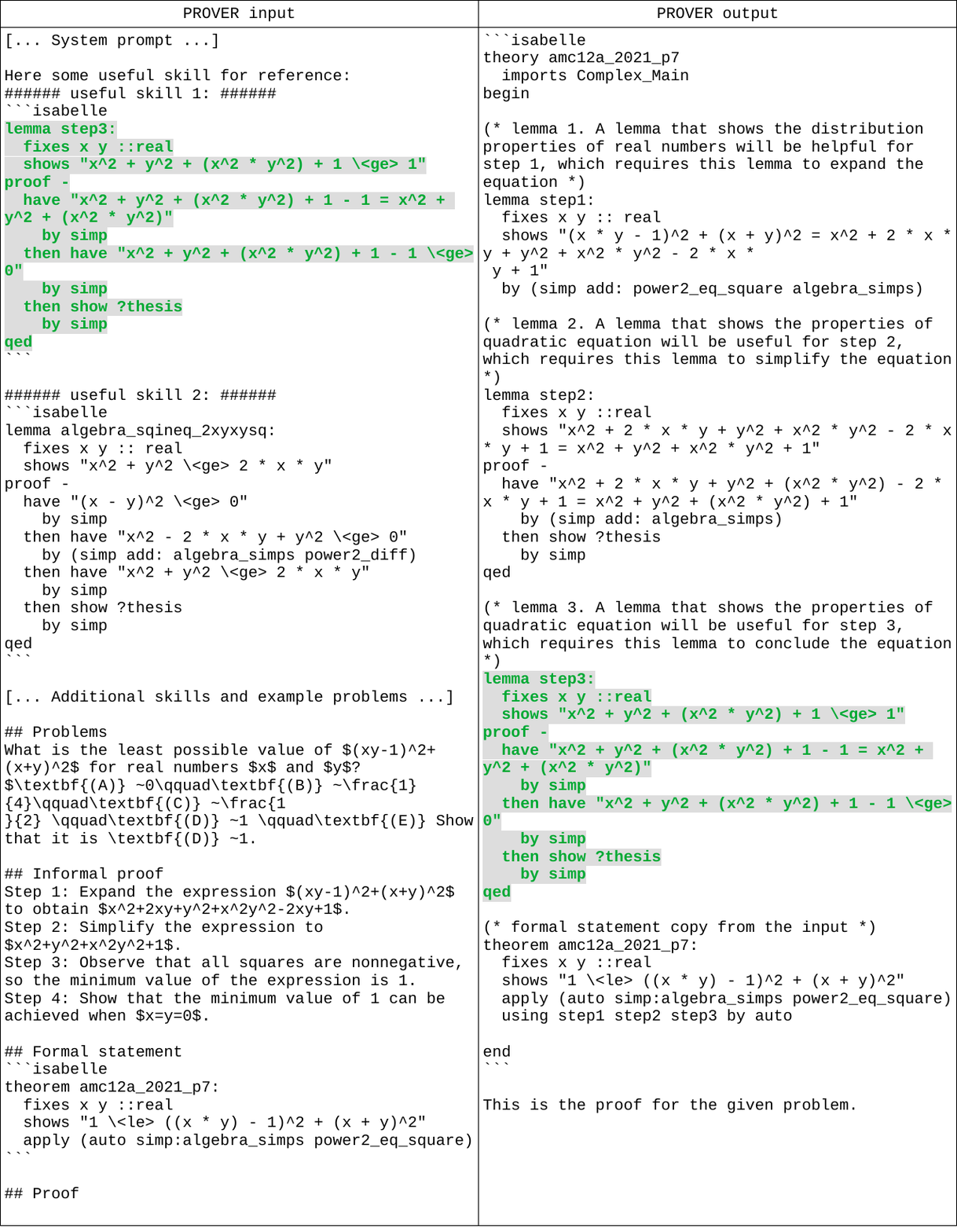}
  \caption{Example of verbatim reuse by the \lp. The input lemma is reproduced exactly in the \prover's output.} \label{fig:app_lp_verbatim}
\end{figure}

Figure \ref{fig:app_lp_verbatim} is an example of verbatim use where an input lemma to the \prover\ is used verbatim in the outputted solution. 

\begin{figure}
  \centering
  \includegraphics[width=\linewidth]{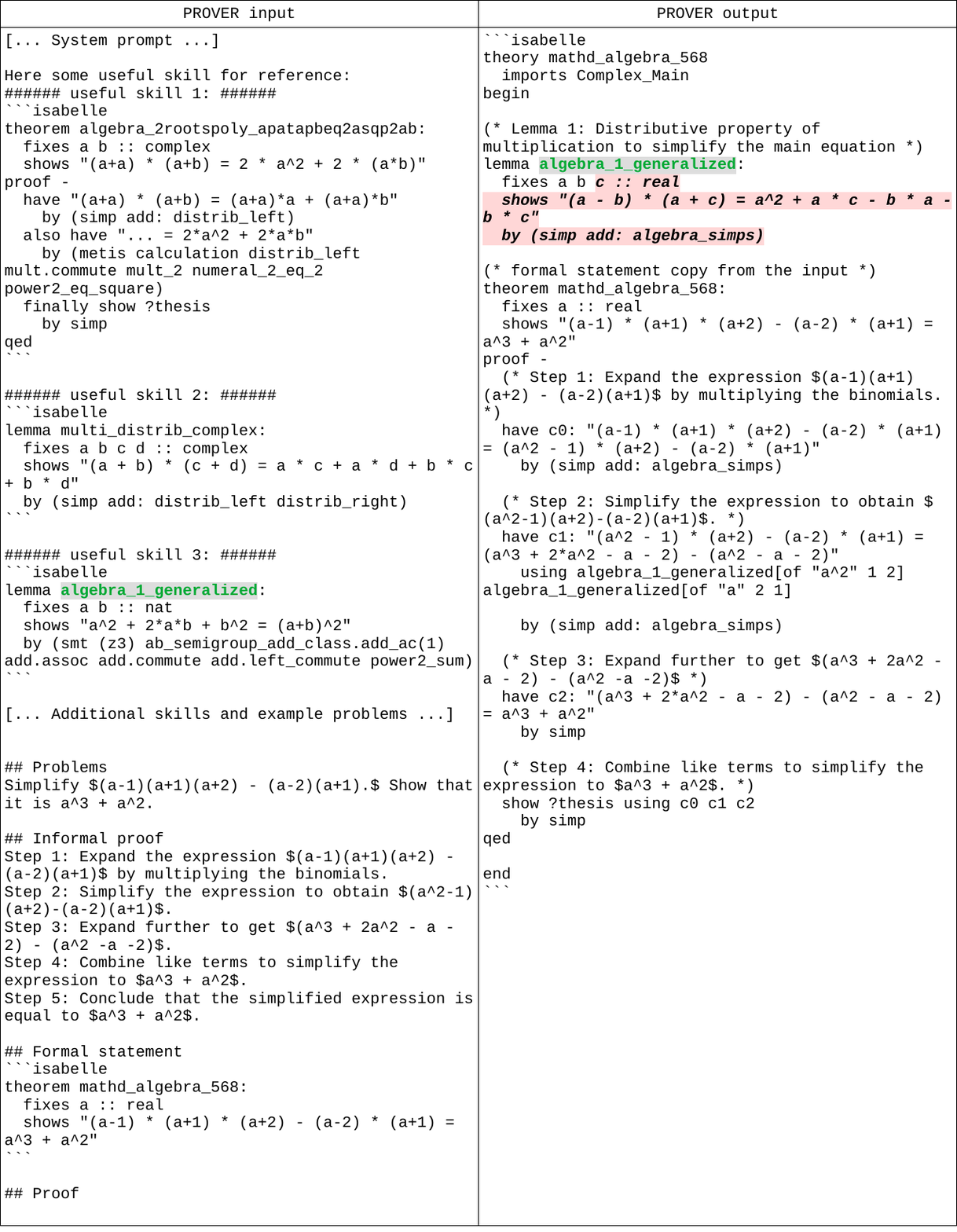}
  \caption{Example of name reuse by the \lp. Only the name of the input lemma needs to be reproduced exactly in the output. In this case, the body of the input lemma has been significantly adjusted. Note Figure \ref{fig:app_lp_verbatim} is also an example of name reuse, as the input lemma's name appears in the solution (in that particular case, along with the rest of the lemma).} \label{fig:app_lp_name}
\end{figure}

In contrast, Figure \ref{fig:app_lp_name} is an example of name use, where the name of the input lemma appears in the solution. In this case, the contents of the lemma are similar, but have significant differences. Note that an instance of verbatim use would, necessarily, also be an instance of name use.

A lemma is reused $N$ times if it is used $N+1$ times -- i.e., if the lemma is used in $N+1$ solutions.

\section{\lp\ Solutions not Found by Reuse-Free Ablation}\label{app:lp_baseline_proofs}

\begin{figure}
  \centering
  \includegraphics[width=\linewidth]{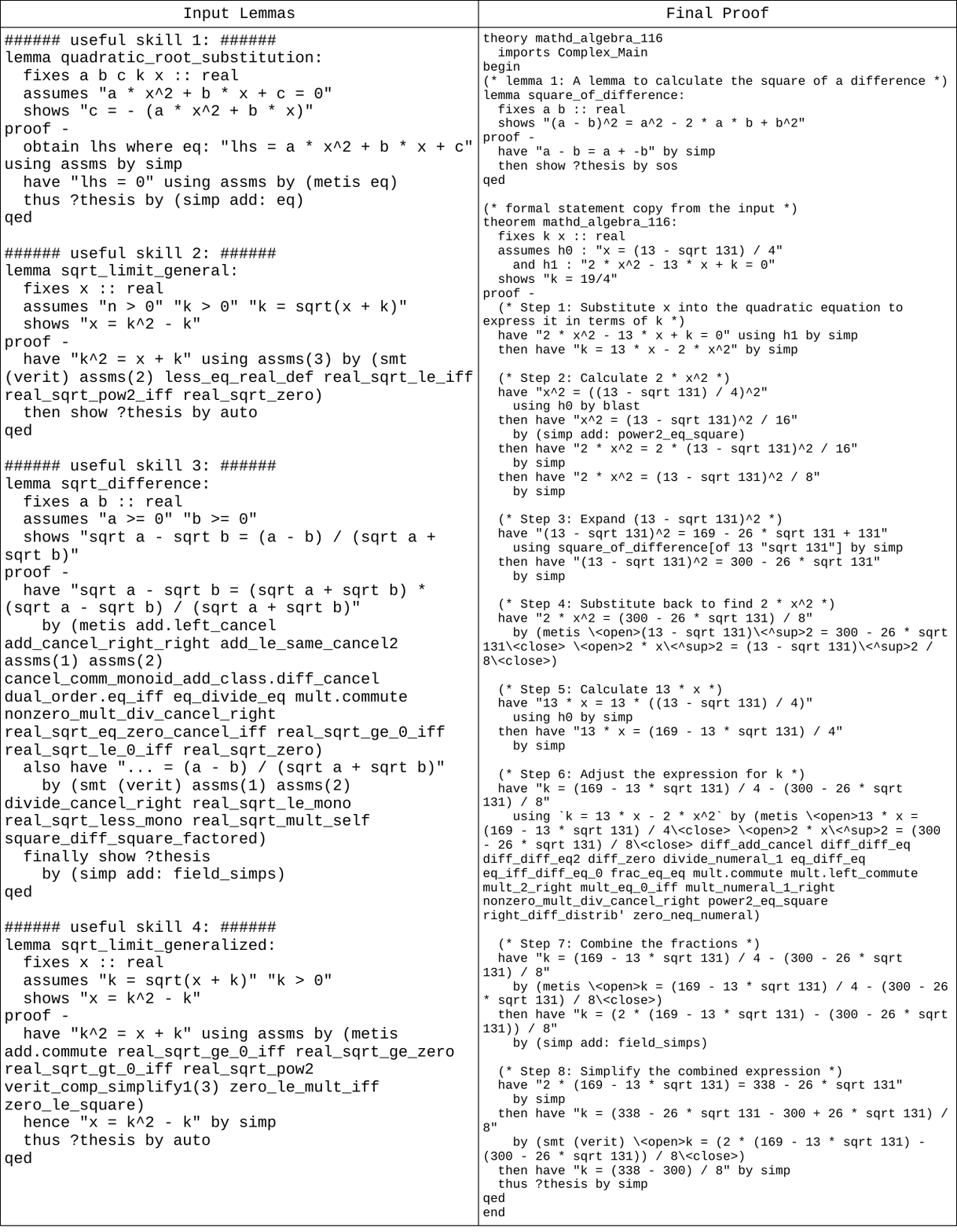}
  \caption{\lp\ input lemmas (left) and found proof (right). The proof proves that $\forall k \in \mathbb{R}:$ if 
	 $x = (13 - \sqrt{131}) / 4$ 
	and $2x^2 - 13x + k = 0$  
	then $k = 19/4$. See Figure \ref{fig:app_lp_mathd_latex} for a typeset approximation, and commentary of \lp's use (and failure to use) the input lemmas.} \label{fig:app_lp_mathd}
\end{figure}

\begin{figure}
  \centering
  \includegraphics[width=\linewidth]{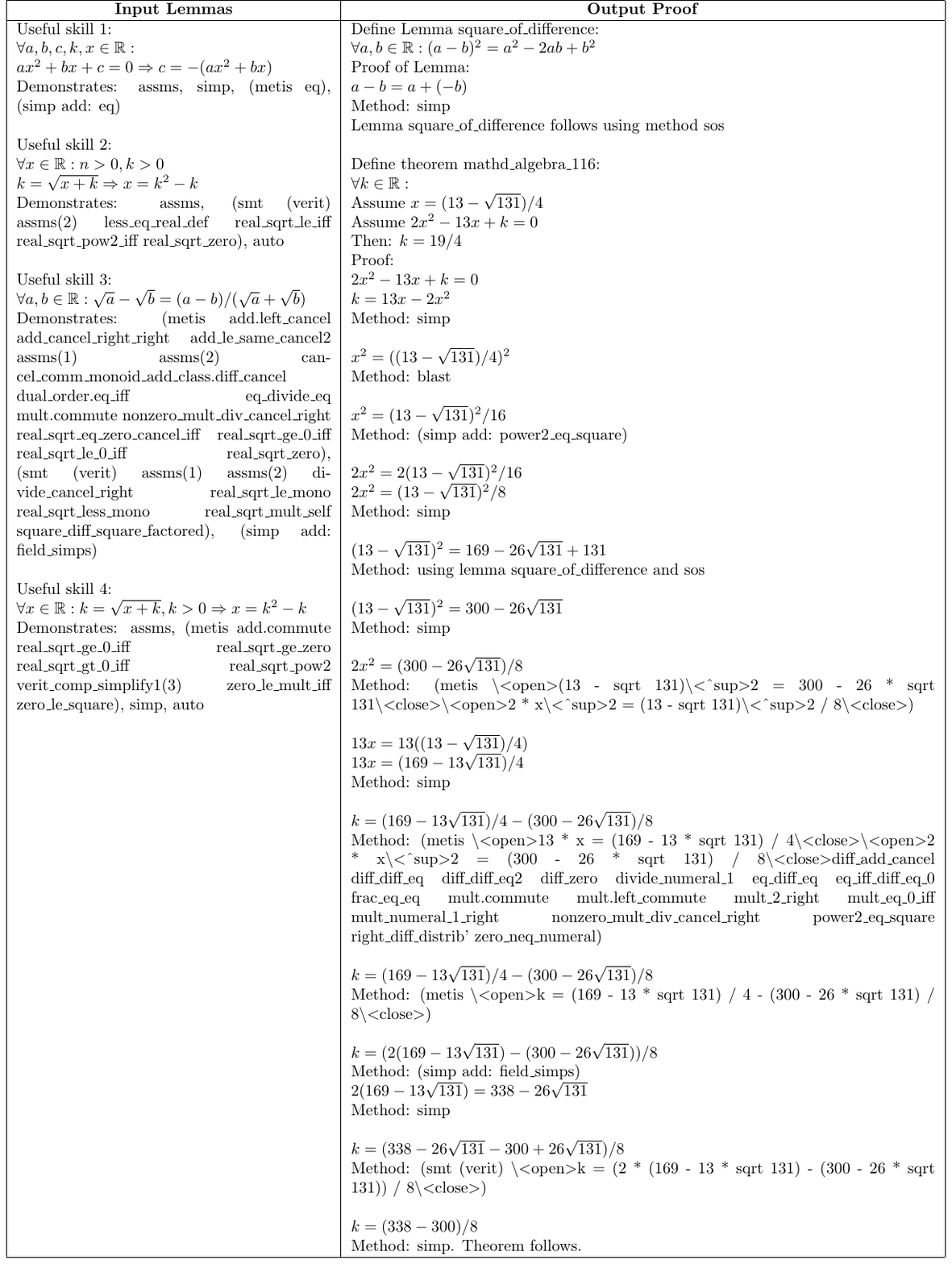}
  \caption{A typset approximation of \lp\ input lemmas (left) and found proof (right). The proof proves that $\forall k \in \mathbb{R}:$ if 
	 $x = (13 - \sqrt{131}) / 4$ 
	and $2x^2 - 13x + k = 0$  
	then $k = 19/4$. See Figure \ref{fig:app_lp_mathd} for the original Isabelle lemmas and proof. Note that skill 1 may have been indirectly used in rewriting $2x^2 - 13x + k = 0$ as $k = 13x - 2x^2$. Skills 2, 3 and 4 do not seem to be used directly or indirectly; furthermore skills 2 and 4 are the same lemma. Their use of smt and metis may have encouraged the prover to use these same tools. } \label{fig:app_lp_mathd_latex}
\end{figure}

\begin{figure}
  \centering
  \includegraphics[width=\linewidth,trim={0 3cm 0 0},clip]{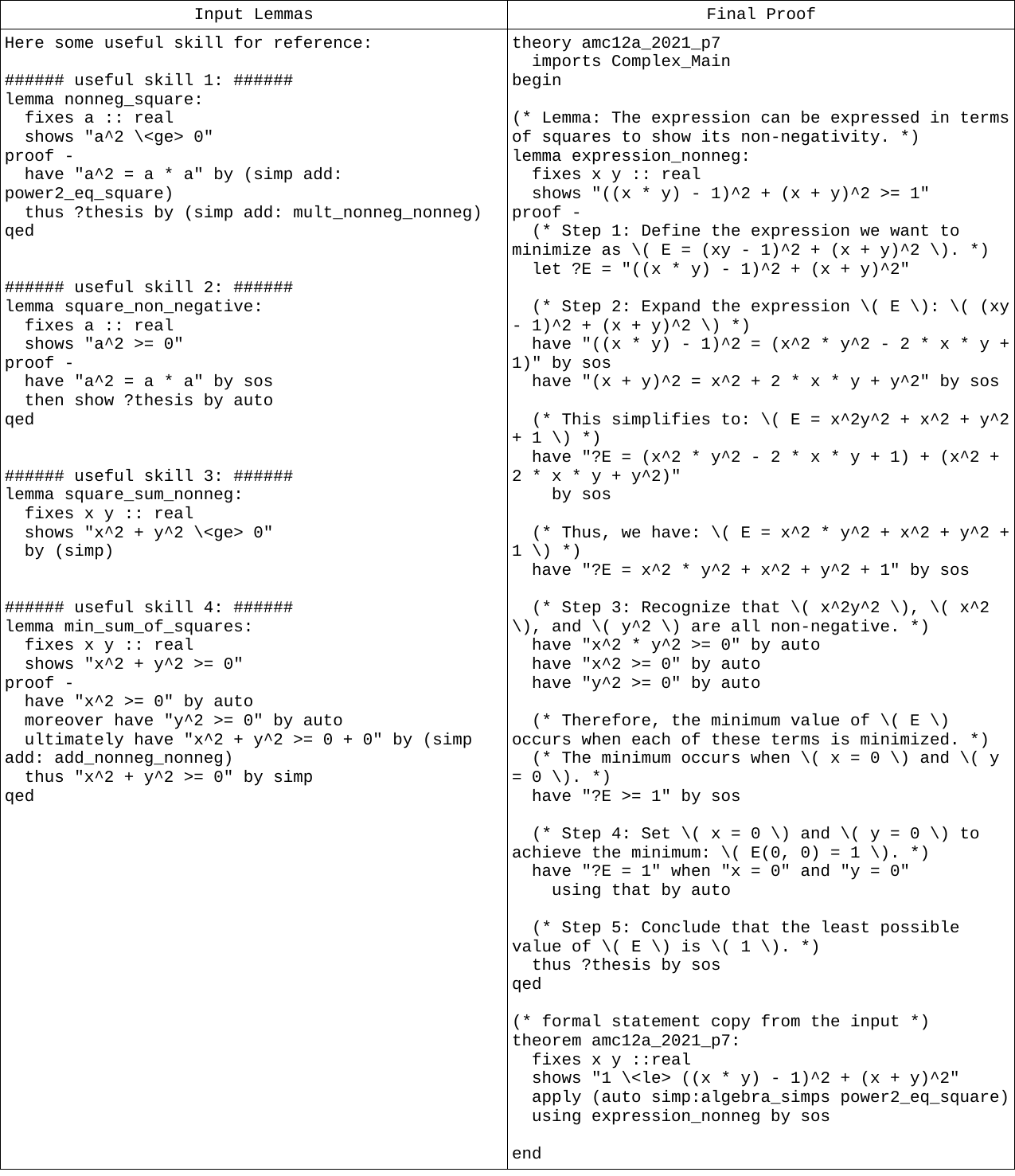}
  \caption{\lp\ input lemmas (left) and found proof (right). The proof proves that $\forall x,y \in \mathbb{R}: 1 \leq (xy - 1)^2 + (x + y)^2$.  See Figure \ref{fig:app_lp_amc_latex} for a typeset approximation, and commentary of \lp's use (and failure to use) the input lemmas.} \label{fig:app_lp_amc}
\end{figure}

\begin{figure}
  \centering
  \includegraphics[width=\linewidth,trim={0 14cm 0 0},clip]{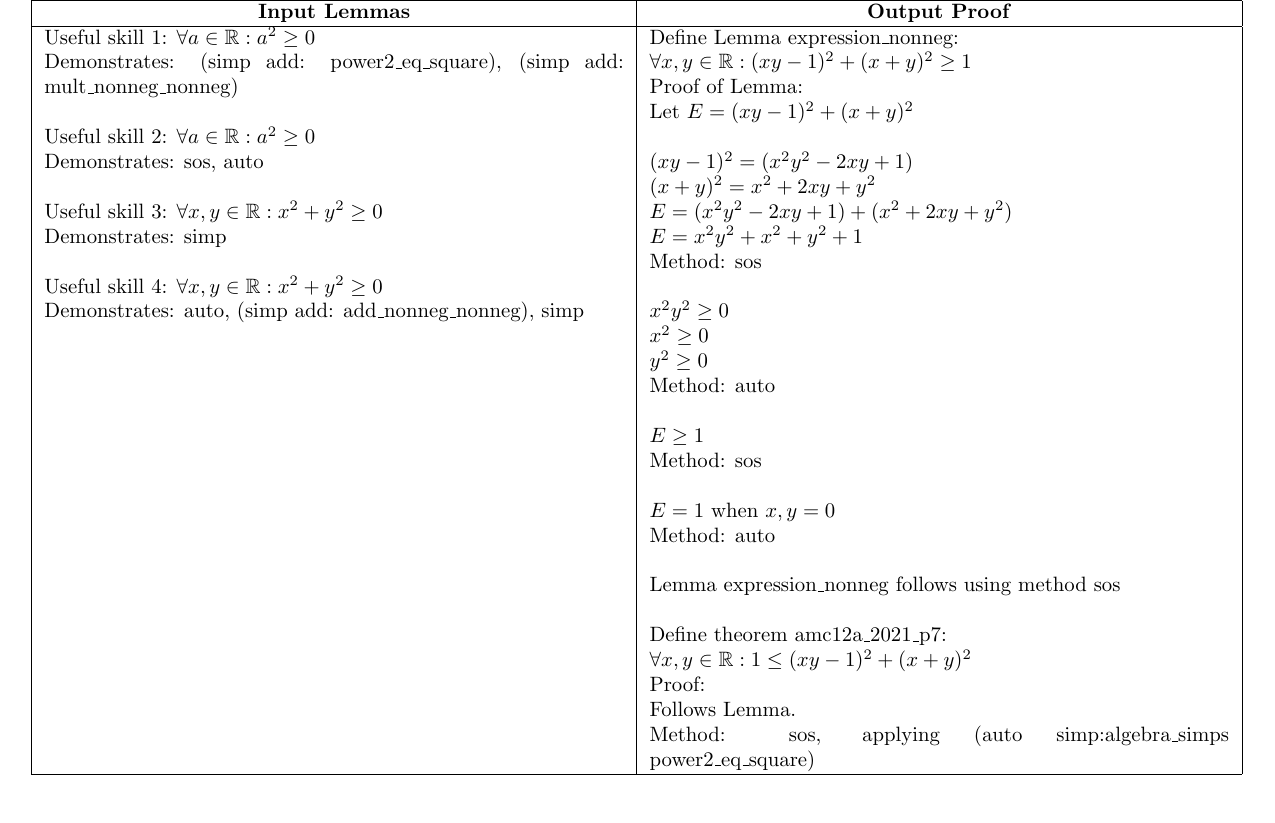}
  \caption{Typeset approximation of \lp\ input lemmas (left) and found proof (right). See Figure \ref{fig:app_lp_amc} for the original Isabelle lemmas and proof. The proof proves that $\forall x,y \in \mathbb{R}: 1 \leq (xy - 1)^2 + (x + y)^2$.  Skills 1 and 2 are the same; the fact that $x^2 \geq 0$ is used, though the exact proof differs from the lemmas. Skills 3 \& 4 are also the same, though they do not seem to be used. } \label{fig:app_lp_amc_latex}
\end{figure}

We performed two runs of the original model, in both cases it outperformed the ablation by solving one additional problem. We present the found proofs and input lemmas in Figures \ref{fig:app_lp_mathd} and \ref{fig:app_lp_amc}. For improved legibility, we also provide a typeset approximation in Figures \ref{fig:app_lp_mathd_latex} and \ref{fig:app_lp_amc_latex}. In addition to the observations in the main paper, it should be noted that there is redundancy among the retrieved lemmas -- deduplication and retrieval of lemmas remain areas for improvement.

\section{\trove\ MATH reproduction}

See table \ref{app:tab_trove_reprod} for the best-of-five accuracies reported by TroVE, and achieved by our reproduction of their results.

\begin{table}
  \caption{\trove\ performance on MATH. For comparison with \citet{trove}, all reported numbers are best over 5 trials. Variation between trials arises from the stochastic sampling of the underlying LLM.} \label{app:tab_trove_reprod}
  \centering
  \begin{tabular}{lllll}
    \toprule
    & \multicolumn{4}{c}{Best-of-5 accuracy on MATH test split}                   \\
    \cmidrule(r){2-5}
    Model     & count     & geo &  inte & num \\
    \midrule
    
    \trove, Reported & \textbf{0.26} & \textbf{0.08} & 0.11 & 0.25     \\
    \trove\, Reproduced (ours)    &  0.24 & 0.06 & 0.13 & 0.27 \\
    \trove, Reported \tcreate-only ablation & 0.14  & 0.06 & 0.05 & 0.16     \\
    No Reuse Ablation (ours)    & 0.25 & 0.05 & \textbf{0.15} & \textbf{0.31}  \\

    \bottomrule
  \end{tabular}
\end{table}

\section{\lp\ Hyperparameters and Experiment Details}\label{app:hyper_lp}

At the time of publication, the \lp\ logs released by \citet{lego} and used in our analysis are available at \url{https://github.com/wiio12/LEGO-Prover/blob/357672c7751cd0c84aff6bf72a3d1bf97614e81d/result/lego_result.zip}.

\lp\ is built on OpenAI's GPT-3.5-Turbo and the 2022 release of the Isabelle proof assistant, specifically using its abilities as a proof verifier. Note that due to the deprecation of the LLMs originally used by \lp\ (\texttt{gpt-3.5-turbo-0301}, \texttt{gpt-3.5-turbo-0613}, \texttt{gpt-3.5-turbo-16k}, \texttt{gpt-3.5-turbo-16k-0613}, \texttt{gpt-3.5-turbo-16k}, \texttt{gpt-3.5-turbo-16k-0613}), we upgrade the underlying LLM from GPT-3.5-Turbo to GPT-4o-mini.

We use the default \lp\ hyperparameters, except for the number of retry attempts which, following \citet{lego}'s ablations, we reduce to 50. See Table \ref{tab:lp_hyper} for details.

\begin{table}
  \caption{\lp\ hyperparameters}
  \label{tab:lp_hyper}
  \centering
  \begin{tabular}{ll}
    \toprule
    Hyperparameter     & value \\
    \midrule
    Solution attempts per problem (num\_attempts) & 50 \\
    Number of \prover\ processes (num\_prover) & 3 \\
    Number of \evolver\ processes (num\_evolver) & 8 \\
    Temperature (temperature) & 0.7 \\
    \bottomrule
  \end{tabular}
\end{table}

Note that the \lp\ is initialized with a seed library of tools, and our ablation retains this initialization. The core claim we aim to disprove is that the model's performance gains predominantly come from reusable lemmas, and our ablation prevents any cross-task reuse.

The specific 12 problems chosen uniformly at random for our ablation study are: 
aime\_1991\_p6.json,
algebra\_2varlineareq\_xpeeq7\_2xpeeq3\_eeq11\_xeqn4.json,
amc12a\_2008\_p15.json,
amc12a\_2013\_p8.json,
amc12a\_2021\_p7.json,
amc12b\_2002\_p3.json,
amc12b\_2003\_p9.json,
mathd\_algebra\_31.json,
mathd\_algebra\_109.json,
mathd\_algebra\_116.json,
mathd\_numbertheory\_149.json, and 
numbertheory\_sqmod4in01d.json

Note that \lp\ requires both the problem statement, and an informal natural language proof for conversion. We use the same human-generated informal proofs as \citet{lego}. The authors bundled said informal proofs inside of the miniF2F .json files listed above, available for download from \url{https://github.com/wiio12/LEGO-Prover/tree/357672c7751cd0c84aff6bf72a3d1bf97614e81d/data/full_data/valid} at the time of publication.

Note that the mean and standard deviation in Figure \ref{fig:lp_ablat} are calculated using Python 3.8.9, numpy 1.22.2, \texttt{numpy.mean()} and \texttt{numpy.std()}.

Our experiments were run on an internal cluster, running one trial at a time. Each trial used 180 GB of RAM, 50 CPU cores, OpenAI credits, and ran within 24 hours. We upper bound the total compute time required to run our \lp\ experiments at 96 hours. The full project required more compute than the experiments reported as one trial failed due to an out-of-memory error. Based on \citet{lego}'s estimate of \$300 per trial, we estimate the cost in OpenAI credits of our experiments to be \$7.38 per trial as we run half the number of attempts and one twentieth the number of questions. Under this estimate, the  total cost of all our experiments is $\sim$\$30.

Our code is modified from the released \lp\ code base, available at \url{https://github.com/wiio12/LEGO-Prover} \citep{trove}, released under an \href{https://github.com/wiio12/LEGO-Prover/tree/master?tab=MIT-1-ov-file#readme}{MIT License}. Evaluation is done using the miniF2F \cite{minif2f} dataset, available at \url{https://github.com/openai/miniF2F/tree/main}, which was released under the \href{https://github.com/openai/miniF2F/blob/main/isabelle/LICENSE}{Apache License Version 2.0}.

Our code is documented and released, alongside the generated \lp\ logs. It is a minor modification to the existing code base, and there is no training stage or new limitations. The code is released under the same license as the parent repository.

\section{\trove\ Hyperparameters and Experiment Details}\label{app:hyper_trove}

\trove\ uses \texttt{CodeLlama-7b-Instruct-hf} \citep{DBLP:journals/corr/abs-2308-12950} interacting with the Python3 interpreter. We use the hyperparameters specified in the paper, outlined in Table \ref{tab:trove_hyper}. The same hyperparameters are used for the ablation, and our reproduction of baseline TroVE. 

\begin{table}
  \caption{\trove\ hyperparameters}
  \label{tab:trove_hyper}
  \centering
  \begin{tabular}{ll}
    \toprule
    Hyperparameter     & value \\
    \midrule
    Library trim frequency (trim\_steps) & 500 \\
    Solution execution timeout in seconds (exec\_timeout) & 100 \\
    top-p (top\_p) & 0.95 \\
    Samples per prompt (num\_return\_sequences) & 5 \\
    Temperature (temperature) & 0.6 \\ 
    Max decode length (max\_new\_tokens) & 512\\
    \bottomrule
  \end{tabular}
\end{table}

The mean and standard deviation of our 5 experiment runs are reported in Table \ref{tab:trove_ablat_details}. They are calculated using Python 3.8.9, numpy 1.22.2, \texttt{numpy.mean()} and \texttt{numpy.std()}. The 2-sided t-test reported the same table is performed using the same version of Python, scipy 1.8.1, \texttt{scipy.stats.ttest\_ind()}, with the settings \texttt{equal\_var=False} and \texttt{alternative='less'}.  %

Our experiments were run on an internal cluster, running up to 4 trials at once. Each trial used 1 Nvidia A40 GPU, 64 GB of RAM, 16 CPU cores, and ran within 12 hours. Smaller datasets completed more quickly. We upper bound the total compute time required to run our \trove\ experiments at 480 hours. The full project required more compute than the experiments reported as we also tried running \trove\ with quantized CodeLlama, CodeLlama 13B and 70B, and GPT-4o-mini.

Our code is modified from the released \trove\ code base, available at \url{https://github.com/zorazrw/trove} \citep{trove}, which was released under the \href{https://github.com/zorazrw/trove/tree/main?tab=CC-BY-SA-4.0-1-ov-file}{CC-BY-SA-4.0 license}. Evaluation is done using the MATH \cite{hendrycksmath2021} dataset, available at \url{https://github.com/hendrycks/math}, which was released under an \href{https://github.com/hendrycks/math?tab=MIT-1-ov-file}{MIT License}.

Our code is documented and released, alongside the generated \trove\ logs. It is a minor modification to the existing code base, and there is no training stage or new limitations. The code is released under the same license as the parent repository.

\subsection{Additional \trove\ experiments}

We also ran baseline \trove\ using the larger CodeLlama 13B model, and found similar results with very little direct function use. The key difference with the 7B model was that a single function was learned for the geometry split, but it was never reused in a correct solution. 

We also attempted to run baseline \trove\ using the 70B model, however we discarded the results as the LLM's ethical safeguards were frequently tripped (e.g., giving reasons such as ``it is not appropriate or ethical to provide assistance with academic assignments or graded exercises'').

\newpage
\section*{NeurIPS Paper Checklist}

\begin{enumerate}

\item {\bf Claims}
    \item[] Question: Do the main claims made in the abstract and introduction accurately reflect the paper's contributions and scope?
    \item[] Answer: \answerYes{} %
    \item[] Justification: We analyze \lp\ logs and ablate the model in Section \ref{sec:lp}, and we analyze the \trove\ logs and ablate the model in Section \ref{sec:trove}. In both cases we find little direct reuse, and our ablation performs similarly.
    \item[] Guidelines:
    \begin{itemize}
        \item The answer NA means that the abstract and introduction do not include the claims made in the paper.
        \item The abstract and/or introduction should clearly state the claims made, including the contributions made in the paper and important assumptions and limitations. A No or NA answer to this question will not be perceived well by the reviewers. 
        \item The claims made should match theoretical and experimental results, and reflect how much the results can be expected to generalize to other settings. 
        \item It is fine to include aspirational goals as motivation as long as it is clear that these goals are not attained by the paper. 
    \end{itemize}

\item {\bf Limitations}
    \item[] Question: Does the paper discuss the limitations of the work performed by the authors?
    \item[] Answer: \answerYes{} %
    \item[] Justification: See Section \ref{sec:lims}. Primary limitations are scope (2 models and 2 datasets), and resource constraints on the ablations.
    \item[] Guidelines:
    \begin{itemize}
        \item The answer NA means that the paper has no limitation while the answer No means that the paper has limitations, but those are not discussed in the paper. 
        \item The authors are encouraged to create a separate "Limitations" section in their paper.
        \item The paper should point out any strong assumptions and how robust the results are to violations of these assumptions (e.g., independence assumptions, noiseless settings, model well-specification, asymptotic approximations only holding locally). The authors should reflect on how these assumptions might be violated in practice and what the implications would be.
        \item The authors should reflect on the scope of the claims made, e.g., if the approach was only tested on a few datasets or with a few runs. In general, empirical results often depend on implicit assumptions, which should be articulated.
        \item The authors should reflect on the factors that influence the performance of the approach. For example, a facial recognition algorithm may perform poorly when image resolution is low or images are taken in low lighting. Or a speech-to-text system might not be used reliably to provide closed captions for online lectures because it fails to handle technical jargon.
        \item The authors should discuss the computational efficiency of the proposed algorithms and how they scale with dataset size.
        \item If applicable, the authors should discuss possible limitations of their approach to address problems of privacy and fairness.
        \item While the authors might fear that complete honesty about limitations might be used by reviewers as grounds for rejection, a worse outcome might be that reviewers discover limitations that aren't acknowledged in the paper. The authors should use their best judgment and recognize that individual actions in favor of transparency play an important role in developing norms that preserve the integrity of the community. Reviewers will be specifically instructed to not penalize honesty concerning limitations.
    \end{itemize}

\item {\bf Theory Assumptions and Proofs}
    \item[] Question: For each theoretical result, does the paper provide the full set of assumptions and a complete (and correct) proof?
    \item[] Answer: \answerNA{} %
    \item[] Justification: This work is empirical.
    \item[] Guidelines:
    \begin{itemize}
        \item The answer NA means that the paper does not include theoretical results. 
        \item All the theorems, formulas, and proofs in the paper should be numbered and cross-referenced.
        \item All assumptions should be clearly stated or referenced in the statement of any theorems.
        \item The proofs can either appear in the main paper or the supplemental material, but if they appear in the supplemental material, the authors are encouraged to provide a short proof sketch to provide intuition. 
        \item Inversely, any informal proof provided in the core of the paper should be complemented by formal proofs provided in appendix or supplemental material.
        \item Theorems and Lemmas that the proof relies upon should be properly referenced. 
    \end{itemize}

    \item {\bf Experimental Result Reproducibility}
    \item[] Question: Does the paper fully disclose all the information needed to reproduce the main experimental results of the paper to the extent that it affects the main claims and/or conclusions of the paper (regardless of whether the code and data are provided or not)?
    \item[] Answer: \answerYes{} %
    \item[] Justification: Hyperparameters are reported in Appendices \ref{app:hyper_lp} and \ref{app:hyper_trove}, the \trove\ and \lp\ codebases are publicly available as are the MATH and miniF2F datasets, our ablations are described in Sections \ref{sec:lp} and \ref{sec:trove}, and we release our code, logs, and log analysis code. As to the underlying LLMs, \trove\ uses open source CodeLlama, and our \lp\ ablation runs on a much smaller dataset to reduce the OpenAI API costs.
    \item[] Guidelines:
    \begin{itemize}
        \item The answer NA means that the paper does not include experiments.
        \item If the paper includes experiments, a No answer to this question will not be perceived well by the reviewers: Making the paper reproducible is important, regardless of whether the code and data are provided or not.
        \item If the contribution is a dataset and/or model, the authors should describe the steps taken to make their results reproducible or verifiable. 
        \item Depending on the contribution, reproducibility can be accomplished in various ways. For example, if the contribution is a novel architecture, describing the architecture fully might suffice, or if the contribution is a specific model and empirical evaluation, it may be necessary to either make it possible for others to replicate the model with the same dataset, or provide access to the model. In general. releasing code and data is often one good way to accomplish this, but reproducibility can also be provided via detailed instructions for how to replicate the results, access to a hosted model (e.g., in the case of a large language model), releasing of a model checkpoint, or other means that are appropriate to the research performed.
        \item While NeurIPS does not require releasing code, the conference does require all submissions to provide some reasonable avenue for reproducibility, which may depend on the nature of the contribution. For example
        \begin{enumerate}
            \item If the contribution is primarily a new algorithm, the paper should make it clear how to reproduce that algorithm.
            \item If the contribution is primarily a new model architecture, the paper should describe the architecture clearly and fully.
            \item If the contribution is a new model (e.g., a large language model), then there should either be a way to access this model for reproducing the results or a way to reproduce the model (e.g., with an open-source dataset or instructions for how to construct the dataset).
            \item We recognize that reproducibility may be tricky in some cases, in which case authors are welcome to describe the particular way they provide for reproducibility. In the case of closed-source models, it may be that access to the model is limited in some way (e.g., to registered users), but it should be possible for other researchers to have some path to reproducing or verifying the results.
        \end{enumerate}
    \end{itemize}

\item {\bf Open access to data and code}
    \item[] Question: Does the paper provide open access to the data and code, with sufficient instructions to faithfully reproduce the main experimental results, as described in supplemental material?
    \item[] Answer: \answerYes{} %
    \item[] Justification: As explained in the previous question on reproducibility, we release our code along with the logs analyzed. Furthermore, the core \trove\ and \lp\ code bases are already publicly available, and can be easily modified to implement the ablations described.
    \item[] Guidelines:
    \begin{itemize}
        \item The answer NA means that paper does not include experiments requiring code.
        \item Please see the NeurIPS code and data submission guidelines (\url{https://nips.cc/public/guides/CodeSubmissionPolicy}) for more details.
        \item While we encourage the release of code and data, we understand that this might not be possible, so “No” is an acceptable answer. Papers cannot be rejected simply for not including code, unless this is central to the contribution (e.g., for a new open-source benchmark).
        \item The instructions should contain the exact command and environment needed to run to reproduce the results. See the NeurIPS code and data submission guidelines (\url{https://nips.cc/public/guides/CodeSubmissionPolicy}) for more details.
        \item The authors should provide instructions on data access and preparation, including how to access the raw data, preprocessed data, intermediate data, and generated data, etc.
        \item The authors should provide scripts to reproduce all experimental results for the new proposed method and baselines. If only a subset of experiments are reproducible, they should state which ones are omitted from the script and why.
        \item At submission time, to preserve anonymity, the authors should release anonymized versions (if applicable).
        \item Providing as much information as possible in supplemental material (appended to the paper) is recommended, but including URLs to data and code is permitted.
    \end{itemize}

\item {\bf Experimental Setting/Details}
    \item[] Question: Does the paper specify all the training and test details (e.g., data splits, hyperparameters, how they were chosen, type of optimizer, etc.) necessary to understand the results?
    \item[] Answer: \answerYes{} %
    \item[] Justification: Hyperparameters are in Sections \ref{app:hyper_lp} and \ref{app:hyper_trove}, there is no training data, and the \trove\ test set is the same as \citet{trove}, and the \lp\ test set a subset of that used in \citet{lego}. The exact problems used in the subset are listed in the same section as the hyperparameters. 
    \item[] Guidelines:
    \begin{itemize}
        \item The answer NA means that the paper does not include experiments.
        \item The experimental setting should be presented in the core of the paper to a level of detail that is necessary to appreciate the results and make sense of them.
        \item The full details can be provided either with the code, in appendix, or as supplemental material.
    \end{itemize}

\item {\bf Experiment Statistical Significance}
    \item[] Question: Does the paper report error bars suitably and correctly defined or other appropriate information about the statistical significance of the experiments?
    \item[] Answer: \answerYes{} %
    \item[] Justification: For the \lp\ ablation, error regions of 1 standard deviation are displayed in Figure \ref{fig:lp_ablat}, the caption states that the source of variation is the LLM output and race conditions within the system; the method used to compute mean and standard deviation (numpy) is stated in Appendix \ref{app:hyper_lp}. For the \trove\ ablation, we report the mean and standard deviation in Table \ref{tab:trove_ablat_details}. The best-of-five accuracy is reported in the Appendix, Table \ref{app:tab_trove_reprod}) so that our values are comparable to those reported in \citet{trove}. Both tables state that variation arises from sampling from the LLM. The method used to compute mean and standard deviation (numpy) is stated in Appendix \ref{app:hyper_trove}.
    \item[] Guidelines:
    \begin{itemize}
        \item The answer NA means that the paper does not include experiments.
        \item The authors should answer "Yes" if the results are accompanied by error bars, confidence intervals, or statistical significance tests, at least for the experiments that support the main claims of the paper.
        \item The factors of variability that the error bars are capturing should be clearly stated (for example, train/test split, initialization, random drawing of some parameter, or overall run with given experimental conditions).
        \item The method for calculating the error bars should be explained (closed form formula, call to a library function, bootstrap, etc.)
        \item The assumptions made should be given (e.g., Normally distributed errors).
        \item It should be clear whether the error bar is the standard deviation or the standard error of the mean.
        \item It is OK to report 1-sigma error bars, but one should state it. The authors should preferably report a 2-sigma error bar than state that they have a 96\% CI, if the hypothesis of Normality of errors is not verified.
        \item For asymmetric distributions, the authors should be careful not to show in tables or figures symmetric error bars that would yield results that are out of range (e.g. negative error rates).
        \item If error bars are reported in tables or plots, The authors should explain in the text how they were calculated and reference the corresponding figures or tables in the text.
    \end{itemize}

\item {\bf Experiments Compute Resources}
    \item[] Question: For each experiment, does the paper provide sufficient information on the computer resources (type of compute workers, memory, time of execution) needed to reproduce the experiments?
    \item[] Answer: \answerYes{} %
    \item[] Justification: Outlined in Appendix \ref{app:hyper_lp} for the \lp\ experiments, and Appendix \ref{app:hyper_trove} for the \trove\ experiments.
    \item[] Guidelines:
    \begin{itemize}
        \item The answer NA means that the paper does not include experiments.
        \item The paper should indicate the type of compute workers CPU or GPU, internal cluster, or cloud provider, including relevant memory and storage.
        \item The paper should provide the amount of compute required for each of the individual experimental runs as well as estimate the total compute. 
        \item The paper should disclose whether the full research project required more compute than the experiments reported in the paper (e.g., preliminary or failed experiments that didn't make it into the paper). 
    \end{itemize}
    
\item {\bf Code Of Ethics}
    \item[] Question: Does the research conducted in the paper conform, in every respect, with the NeurIPS Code of Ethics \url{https://neurips.cc/public/EthicsGuidelines}?
    \item[] Answer: \answerYes{} %
    \item[] Justification: There are no human subjects, to the best of our knowledge there are no data concerns, or immediate societal impact or harms (the possible future risks from deploying tool-learning systems, and the precautions that should be taken in future research in self-improving systems are outlined in Section \ref{sec:lims}), and to the best of our knowledge our work is reproducible and legal.
    \item[] Guidelines:
    \begin{itemize}
        \item The answer NA means that the authors have not reviewed the NeurIPS Code of Ethics.
        \item If the authors answer No, they should explain the special circumstances that require a deviation from the Code of Ethics.
        \item The authors should make sure to preserve anonymity (e.g., if there is a special consideration due to laws or regulations in their jurisdiction).
    \end{itemize}

\item {\bf Broader Impacts}
    \item[] Question: Does the paper discuss both potential positive societal impacts and negative societal impacts of the work performed?
    \item[] Answer: \answerYes{} %
    \item[] Justification: We do not anticipate any immediate societal impact or harms, but we do discuss the possible future risks from deploying tool-learning systems, and the precautions that should be taken in future research in self-improving systems in Section \ref{sec:lims}.
    \item[] Guidelines:
    \begin{itemize}
        \item The answer NA means that there is no societal impact of the work performed.
        \item If the authors answer NA or No, they should explain why their work has no societal impact or why the paper does not address societal impact.
        \item Examples of negative societal impacts include potential malicious or unintended uses (e.g., disinformation, generating fake profiles, surveillance), fairness considerations (e.g., deployment of technologies that could make decisions that unfairly impact specific groups), privacy considerations, and security considerations.
        \item The conference expects that many papers will be foundational research and not tied to particular applications, let alone deployments. However, if there is a direct path to any negative applications, the authors should point it out. For example, it is legitimate to point out that an improvement in the quality of generative models could be used to generate deepfakes for disinformation. On the other hand, it is not needed to point out that a generic algorithm for optimizing neural networks could enable people to train models that generate Deepfakes faster.
        \item The authors should consider possible harms that could arise when the technology is being used as intended and functioning correctly, harms that could arise when the technology is being used as intended but gives incorrect results, and harms following from (intentional or unintentional) misuse of the technology.
        \item If there are negative societal impacts, the authors could also discuss possible mitigation strategies (e.g., gated release of models, providing defenses in addition to attacks, mechanisms for monitoring misuse, mechanisms to monitor how a system learns from feedback over time, improving the efficiency and accessibility of ML).
    \end{itemize}
    
\item {\bf Safeguards}
    \item[] Question: Does the paper describe safeguards that have been put in place for responsible release of data or models that have a high risk for misuse (e.g., pretrained language models, image generators, or scraped datasets)?
    \item[] Answer: \answerNA{} %
    \item[] Justification: We present ablations of already publicly available models (\lp\ and \trove), neither of which we believe has a higher risk for misuse than the constituent publicly available LLM.
    \item[] Guidelines:
    \begin{itemize}
        \item The answer NA means that the paper poses no such risks.
        \item Released models that have a high risk for misuse or dual-use should be released with necessary safeguards to allow for controlled use of the model, for example by requiring that users adhere to usage guidelines or restrictions to access the model or implementing safety filters. 
        \item Datasets that have been scraped from the Internet could pose safety risks. The authors should describe how they avoided releasing unsafe images.
        \item We recognize that providing effective safeguards is challenging, and many papers do not require this, but we encourage authors to take this into account and make a best faith effort.
    \end{itemize}

\item {\bf Licenses for existing assets}
    \item[] Question: Are the creators or original owners of assets (e.g., code, data, models), used in the paper, properly credited and are the license and terms of use explicitly mentioned and properly respected?
    \item[] Answer: \answerYes{} %
    \item[] Justification: The creators of \trove\ \citep{trove}, \lp\ \citep{lego}, the MATH dataset \citep{hendrycksmath2021}, and miniF2F \citep{minif2f} are all cited in the abstract. The URLs and licenses  are stated in Appendices \ref{app:hyper_lp} and \ref{app:hyper_trove}.
    \item[] Guidelines:
    \begin{itemize}
        \item The answer NA means that the paper does not use existing assets.
        \item The authors should cite the original paper that produced the code package or dataset.
        \item The authors should state which version of the asset is used and, if possible, include a URL.
        \item The name of the license (e.g., CC-BY 4.0) should be included for each asset.
        \item For scraped data from a particular source (e.g., website), the copyright and terms of service of that source should be provided.
        \item If assets are released, the license, copyright information, and terms of use in the package should be provided. For popular datasets, \url{paperswithcode.com/datasets} has curated licenses for some datasets. Their licensing guide can help determine the license of a dataset.
        \item For existing datasets that are re-packaged, both the original license and the license of the derived asset (if it has changed) should be provided.
        \item If this information is not available online, the authors are encouraged to reach out to the asset's creators.
    \end{itemize}

\item {\bf New Assets}
    \item[] Question: Are new assets introduced in the paper well documented and is the documentation provided alongside the assets?
    \item[] Answer: \answerYes{} %
    \item[] Justification: Our code is documented and released, alongside the log files used in our analysis. As new assets are minor modifications to existing code bases with no training or new limitations, we simply state as much in Appendices \ref{app:hyper_lp} and \ref{app:hyper_trove}; the code will be released under the same license as the parent repositories.
    \item[] Guidelines:
    \begin{itemize}
        \item The answer NA means that the paper does not release new assets.
        \item Researchers should communicate the details of the dataset/code/model as part of their submissions via structured templates. This includes details about training, license, limitations, etc. 
        \item The paper should discuss whether and how consent was obtained from people whose asset is used.
        \item At submission time, remember to anonymize your assets (if applicable). You can either create an anonymized URL or include an anonymized zip file.
    \end{itemize}

\item {\bf Crowdsourcing and Research with Human Subjects}
    \item[] Question: For crowdsourcing experiments and research with human subjects, does the paper include the full text of instructions given to participants and screenshots, if applicable, as well as details about compensation (if any)? 
    \item[] Answer: \answerNA{} %
    \item[] Justification: No crowdsourcing or human subjects was done.
    \item[] Guidelines:
    \begin{itemize}
        \item The answer NA means that the paper does not involve crowdsourcing nor research with human subjects.
        \item Including this information in the supplemental material is fine, but if the main contribution of the paper involves human subjects, then as much detail as possible should be included in the main paper. 
        \item According to the NeurIPS Code of Ethics, workers involved in data collection, curation, or other labor should be paid at least the minimum wage in the country of the data collector. 
    \end{itemize}

\item {\bf Institutional Review Board (IRB) Approvals or Equivalent for Research with Human Subjects}
    \item[] Question: Does the paper describe potential risks incurred by study participants, whether such risks were disclosed to the subjects, and whether Institutional Review Board (IRB) approvals (or an equivalent approval/review based on the requirements of your country or institution) were obtained?
    \item[] Answer: \answerNA{} %
    \item[] Justification: There were no human study participants.
    \item[] Guidelines:
    \begin{itemize}
        \item The answer NA means that the paper does not involve crowdsourcing nor research with human subjects.
        \item Depending on the country in which research is conducted, IRB approval (or equivalent) may be required for any human subjects research. If you obtained IRB approval, you should clearly state this in the paper. 
        \item We recognize that the procedures for this may vary significantly between institutions and locations, and we expect authors to adhere to the NeurIPS Code of Ethics and the guidelines for their institution. 
        \item For initial submissions, do not include any information that would break anonymity (if applicable), such as the institution conducting the review.
    \end{itemize}

\end{enumerate}

\end{document}